\documentclass[11pt]{article}
\usepackage[left=1in,right=1in,top=1in,bottom=1in]{geometry}
\usepackage{booktabs}
\usepackage{times}
\usepackage{graphicx}
\usepackage{dsfont}
\usepackage{amsmath}
\usepackage{amsfonts}
\usepackage{amssymb}
\usepackage{authblk}

\begin{document}
\title{Crime Hot-Spot Modeling via Topic Modeling\\ and Relative Density Estimation}
\author[1]{Jonathan Zhou}
\author[1]{Sarah Huestis-Mitchell}
\author[2]{Xiuyuan Cheng}
\author[1]{Yao Xie}

\affil[1]{Georgia Institute of Technology, Atlanta GA, 30332} 
\affil[2]{Duke University, Durham NC, 27708}
\date{}
\maketitle

\begin{abstract}
    We present a method to capture groupings of similar calls and determine their relative spatial distribution from a collection of crime record narratives. We first obtain a topic distribution for each narrative, and then propose a nearest neighbors relative density estimation (kNN-RDE) approach to obtain spatial relative densities per topic. Experiments over a large corpus ($n=475,019$) of narrative documents from the Atlanta Police Department demonstrate the viability of our method in capturing geographic hot-spot trends which call dispatchers do not initially pick up on and which go unnoticed due to conflation with elevated event density in general.
\end{abstract}

\section{Introduction}
\label{sec:introduction}

In a typical emergency call for service event, a 911 dispatcher takes down information from the caller, dispatches an officer, and attaches available location, type, time, and text information to the call record. Eventually, the officer arrives on scene, handles the situation at hand, and then writes up a detailed report of what occurred during the incident, known as an ``\textit{incident narrative}''. Such narratives are extremely informative and contain nuance beyond the set of time-points and event category assigned by a call dispatcher and recorded in a call for service database. However, the narratives are also highly noisy and unstructured as they are often written in haste. In this work, we study an extensive call-for-service data set from the Atlanta Police Department (APD). The Atlanta Police Department (APD) is one of the largest police departments in the southeastern United States, handing well over a million 911 calls per year \cite{hepburn2021}.

In the face of increasing call and patrol volumes, a major problem the police face is the deployment of their limited resources to engage with the community. To help address these challenges, we detail a method to utilize these incident narratives to identify interpretable groupings of similar events and characterize their relative spatial distribution. This enables the department to respond to an identified class of events in aggregate. Our approach not only captures groupings and detects additional geographic hot-spot trends that call dispatchers and the police force may not initially pick up on themselves but also enables the identification of trends that may go unnoticed due to conflation with higher population or elevated event density, which we address through relative density estimation.

Our work is the first to identify the spatial distributions of grouping of police incidents and generate relative event densities of topics based on documents that describe these incidents. We consider topics generated from two clustering algorithms (Latent Dirichlet Allocation \cite{blei2003} and Non-negative Matrix Factorization \cite{lee1999}) which are run on a large corpus of narratives. We then evaluate the consistency of identified topics and model performance based on geographical hot-spot detection. Finally, we analyze the spatial distributions of the clusters of incidents created by the topics.

\subsection{Relation to Prior Work}
Prior work has explored police event topic identification in the field of criminology \cite{neu2020}, such as an influential book by Brantingham \cite{brantingham1984patterns} detailing crime trends across countries. These studies have remained qualitative and do not use automatic topic discovery. Automatic narrative topic identification has been undertaken using data from the Los Angeles Police Department (LAPD) \cite{pandey2017, kuang2017}, but does not characterize a continuous spatial distribution of each topic. These two works were the first to apply Non-Negative Matrix factorization (NMF) to police topic modeling. \cite{pandey2017} focused on NMF factorization and the Gini index as the key metric to measure spatial distribution, which provides a measure of how unequally a topic is distributed across a grid of places in a city. They examined cosine similarities between 7 learned topics and found insights not apparent in manually assigned incident types. In particular, they calculate the Gini index for seven topics in each square of a discrete 100 $\times$ 100 grid. \cite{kuang2017} extended this using a subset of the same dataset but did not provide a spatial distribution and only compared between the topics. They compared LDA and NMF topics using coherence and Gini index measures, concluding that LDA topics exhibited better metrics and a higher Gini index.

By contrast, we introduce a framework for estimating the relative density of a specific incident type compared to all incidents at a given point. Rather than using methods like Gaussian kernel density or Parzen window to gauge incident type density, we assess the ratio of density at a point for the given type relative to the mean density. This approach cleanly reveals the frequency of occurrences for a certain incident type in a specific area and also yields a continuous two-dimensional relative density estimate at any point with good computational efficiency. This eliminates the need for additional calculations and enhances hotspot characterization. The use of relative density is motivated since crime or other police events may be concentrated simply due to there being greater population or the concentration of other non-crime related phenomena \cite{eck2017}.

Our study marks the first instance of applying topic modeling and density estimation to Atlanta Police Department data. \cite{hepburn2021} outlines ongoing research about Atlanta 911 calls, and \cite{baldwin2020} presents an extensive analysis of Atlanta's 911 calls between 2017 and 2020.

\section{Data}
\label{sec:data_sources}
We use a dataset of 475,019 emergency calls collected between January 2013 to May 2017 and provided by the Atlanta Police Department. Each incident includes a textual description and geographical location in coordinate form. Each call is also associated with one of 273 preliminary call types (e.g. ``Theft from Auto'', and ``Vandalism''), assigned by the call dispatcher \cite{sigcodesop}, though these categorizations only represent the dispatcher's knowledge on call arrival.

\subsection{Call Description Preprocessing}
For each description, we construct a bag-of-words model to obtain tractable numeric representations suitable for topic modeling. We first convert all documents to lowercase and remove all special characters and stopwords \cite{nltk}. We then split every description by sentence and tokenize by word such that each description becomes a list of sentences which in turn are lists of words.
\paragraph{Dictionary Generation and Document Vector Generation} Since bag-of-words models do not preserve word order, we compute all word pairs (2-grams) in each sentence across all the documents to preserve additional context. We use both word pairs and words for subsequent analysis. To remove further words/word-pairs that do not contribute to the meaning of each document, we eliminate words/word-pairs with term frequency-inverse document (TF-IDF) values in the bottom 0.2 quantile \cite{Ramos2003UsingTT}. The above procedure yields a dictionary of 43,676 words and word-pairs. Finally, we convert each tokenized call description into a 43,676 dimensional bag-of-words vector where the $i^{th}$ entry is the occurrence count for the $i^{th}$ word/word-pair in the description.

\section{Methodology}
\subsection{Topic Modeling}
For the narrative topic modeling, we evaluate two topic identification algorithms --- Latent Dirichlet Allocation and Non-negative Matrix Factorization.
\label{sec:topic-modeling}

\textbf{Latent Dirichlet Allocation (LDA)} is a generative probabilistic model for topic extraction from a corpus \cite{blei2003, blei2012}. LDA supposes that there $t$ number of latent topics that underlie a corpus, and topics are represented by a distribution over words. Each document may be then thought of as a sampled mixture of the topics. 

LDA assumes the following generative process underlies document creation. Consider a corpus $D$ consisting of $n$ documents having lengths $l_i$. First suppose there is $\theta_i \sim \mathrm{Dir}(\alpha)$, for $1 \leq i < n$, where $\mathrm{Dir}(\alpha)$ is a Dirichlet distribution with symmetric parameter $\alpha$. Subsequently, consider $\varphi \sim \mathrm{Dir}(\beta)$ across $1 \leq k < t$ , where $t$ is the number the topics. Then the $j$th term of the $i$th document in the corpus $D$ is generated by first selecting a topic $z_{i,j} \sim \mathrm{Multinomial}(\theta_i)$ and then a word $w_{i,j}\sim\mathrm{Multinomial}(\varphi_{z_{i,j}})$. The LDA aims to find parameters $\alpha$ and $\beta$ that maximize the (marginal) log-likelihood of the corpus. We interpret the parameters $\mathbf{\theta }_{d=1 \ldots t }$ as the topic distribution in document $i$.

\textbf{Non-negative matrix factorization (NMF)} is a popular non-probabilistic method for topic modeling and dimensionality reduction  \cite{lee1999,xu2003,chen2010,wang2013,gillis2014}. We represent the collection of documents as a matrix $\mathbf{D} \in \mathbb{N}^{m \times n}$ where $m$ is the dictionary size and $n$ is the number of documents. Each column $\mathbf{D}_i\in \mathbb{N}^m$ represents the bag-of-words vector for document $i$.

Our objective is to find two non-negative matrices $\mathbf{W} \in \mathbb{R}_{\geq 0}^{m \times t}$ and $\mathbf{H} \in \mathbb{R}_{\geq 0}^{t \times n}$ such that $\mathbf{D} \approx \mathbf{W}\times \mathbf{H}$. Here $t$ is the number of topics, a hyperparameter chosen before performing the factorization.  The matrix $\mathbf{W}$ is interpreted as a term-to-topic mapping where each term-topic pair is assigned a weight. In a given topic, highly weighted terms are representative of the topic's meaning. The matrix $\mathbf{H}$ represents a topic-to-document mapping, where each column represents the relative weight of each topic within the description (e.g., description 1 is $20\text{\%}$ about Topic 1 and $80 \text{\%}$ about topic 2).

To compute $\mathbf{W}$ and $\mathbf{H}$, we use the \textit{online-NMF with outliers algorithm} to solve the following minimization problem iteratively and in batches via projected gradient descent \cite{7676413}

$$\min_{\mathbf{W} \in \mathbb{R}_{\geq 0}^{m \times t}, \{\mathbf{h}_i\}_{i=1}^n \geq 0 } \sum_{i=1}^{n} \frac{1}{n} \|\mathbf{d}_i - \mathbf{W}\mathbf{h}_i\|_2^2.$$

\subsection{Topic Relative Density Estimation (RDE) with kNN}
\label{sec:knn-relative-density}

\begin{figure}
    \centering
    \includegraphics[width=0.4\linewidth]{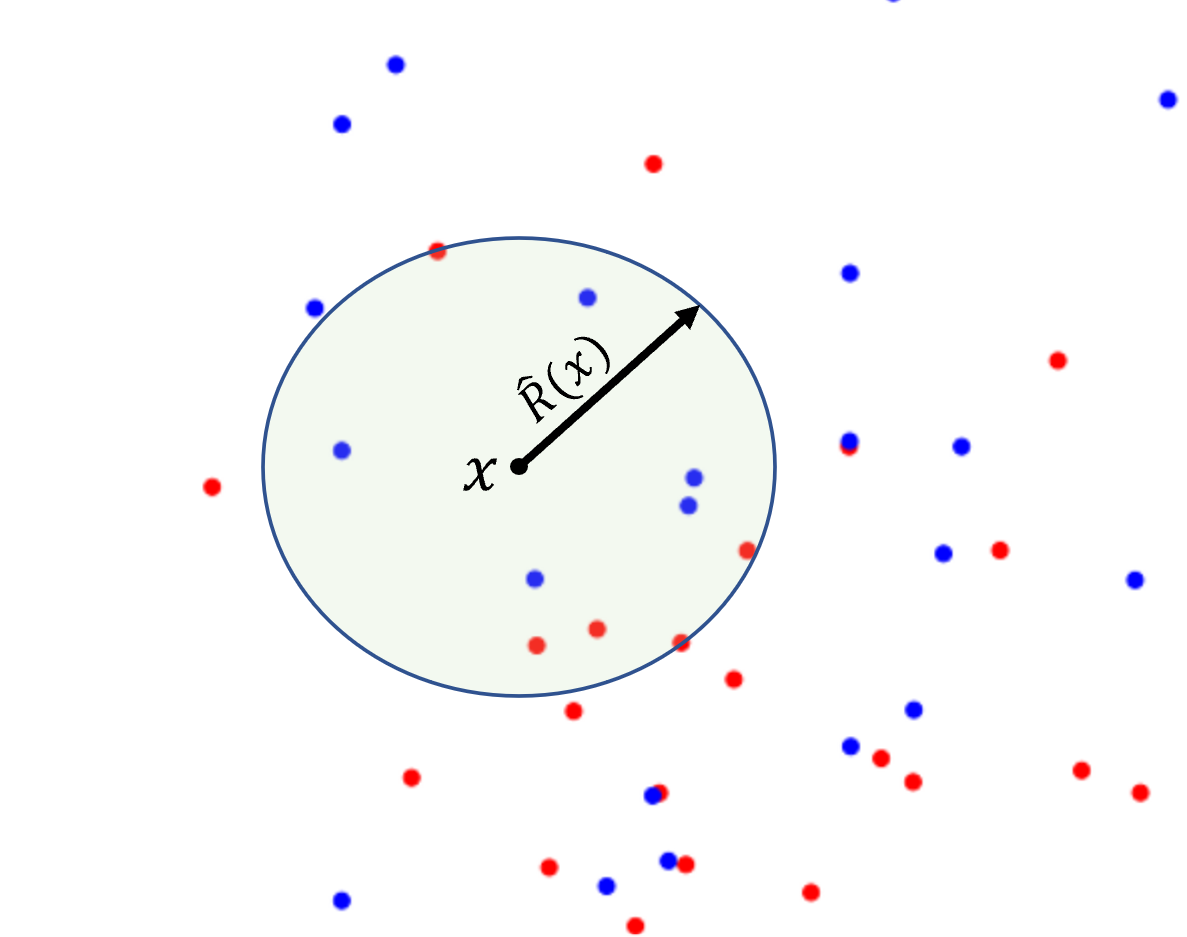} 
    \caption{A sample kNN RDE calculation with $k=10$ and two topics (blue and red). In this case, there are no mixtures of categories in the data points. We compute the kNN RDE at point $x$ by finding the smallest ball of radius $\hat{R}(x)$, such that the number of events within the ball is $k$. Across the topics in the ball,  we calculate the weight of the topic by taking the mean across the weights for each topic for each event in the ball (in this case, 0.5 for blue and red, respectively). Lastly, we divide by the overall average weight of blue events (0.48) and red events (0.52), resulting in a relative density of $\hat{g}_{\rm blue}=\frac{0.5}{0.48}=1.04$ and $\hat{g}_{\rm red}=\frac{0.5}{0.52}=0.96$.}
    \label{fig:knnrde-ex}
\end{figure}

After identifying topics using the procedures above, we empirically estimate the relative density of topic $i \in [t]$ with respect to the overall density of events using a k-nearest Neighbors (kNN) approach to obtain point-wise density. We then use a Gaussian kernel to convert the pointwise topic densities topic into a continuous density across physical space.

More formally, given event data $Z=\{(x_i,c_i)\}_{i=1}^n$ where $x_i \in \mathbb{R}^2$ represents a geographical location and $c_i \in \mathbb{R}_{\geq 0 }^{t}$ is a probability vector representing the relative weighting of $t$ topics contains within the event (e.g., a topic mixture uncovered by NMF or LDA or the call type assigned by the 911 dispatcher). We define $f(x):\mathbb{R}^2\rightarrow[0,1]$ as the probability density of events, such that $\int_{\mathbb{R}^2} f(x) \mathrm{d}x=1$ and $\int_\Omega f(x) \mathrm{d}x$ represents the probability of an event occurring in spatial region $\Omega \in \mathbb{R}^2$. Then, for each topic $m$ we likewise define the probability density function $f_m(x) = \mathbb{P}[x_i =x |m_i=m]$, such that $\int_\Omega f_m(x) \mathrm{d}x$ represents the probability of occurrence of an event of $m$ in area $\Omega$. Our objective is to estimate the relative density ratio $r_m(x)=\frac{f_m(x)}{f(x)}$, which represents the relative proportion in the density of events of topic $m$ when compared to the overall density of events. This allows us to identify specific regions that are more or less incident-crowded as compared to the overall incident rate, with respect to incident type.

We wish to obtain the following $f_1,...,f_t$, probability densities for each of the $t$ topics, such that:
\begin{align*}
    f &= \sum\limits_{m=1}^t \rho_mf_m \text{ where } \sum\limits_{m=1}^t \rho_m=1 \text{ and }
    \forall \ m \in [t],  \rho_t >0
\end{align*}
i.e., the event data $Z=(x_i,c_i)_{i=1}^n$ is sampled i.i.d. from $f$.

We can achieve an estimate for $r_m$ by using a k-Nearest Neighbors approach. Let $n=|Z|$ and $n_m = \sum_{i=1}^n c_{i,m}$, the total across all incidents of the proportion assigned to topic $m$. We first let
\begin{equation*}
    \hat{\rho}_m:=\frac{n_m}{n} \text{ so that } \sum\limits_{m=1}^{M} {\hat{\rho}_m}  =1.
\end{equation*}

To estimate $r_m$ at some point in $x \in \mathbb{R}^2$, we choose an appropriate $k$ neighbor neighbors, and compute the radius of the smallest ball such that the ball contains the $k^{\text{th}}$ nearest neighbors of $x$:
\begin{equation*}
    \hat{R}(x)= \min \{r \in \mathbb{R} : k \leq \sum \limits_{i=1}^n \mathds{1} \{||x-z_i|| < r\} \}.
\end{equation*} 
Once we have obtained $\hat{R}(x)$ we can then compute the total contribution of topic $m$ across the $k$ points, $\mathbb{N}_m(x)$, and $\hat{g}_m(x)$, the empirical relative density of topic $m$ at point $x$. 
\begin{equation*}
    \mathbb{N}_m := \sum\limits_{i=1}^n c_{i,m} \mathds{1} \{||x-x_i||\leq \hat{R}(x) \} \text{ , }   \hat{g}_m(x) :=\frac{\mathbb{N}_m(x)/k}{\hat{\rho}_m}.
\end{equation*}

Then we use $\mathbb{N}_m(x)/k$ as a posterior estimate:
\begin{align*}
    \frac{\mathbb{N}_m(x)}{k} &\approx
    \frac{f_m(x)\rho_m}{f(x)},
\end{align*}

which we use to compute $\hat{g}_m$, the estimate for $r_m(x)$, as follows. Figure \ref{fig:knnrde-ex} provides a detailed computational example. 
$$ \hat{g}_m =\frac{\frac{\mathbb{N}_m}{k}}{\rho_m}\approx \frac{\frac{f_m(x)\rho_m}{f(x)}}{\rho_m}
    =\frac{f_m(x)}{f(x)}.
$$

\section{Results}
\label{sec:results}

\subsection{Topic Modeling}

\begin{figure}
    \centering
    \includegraphics[width=0.8\linewidth]{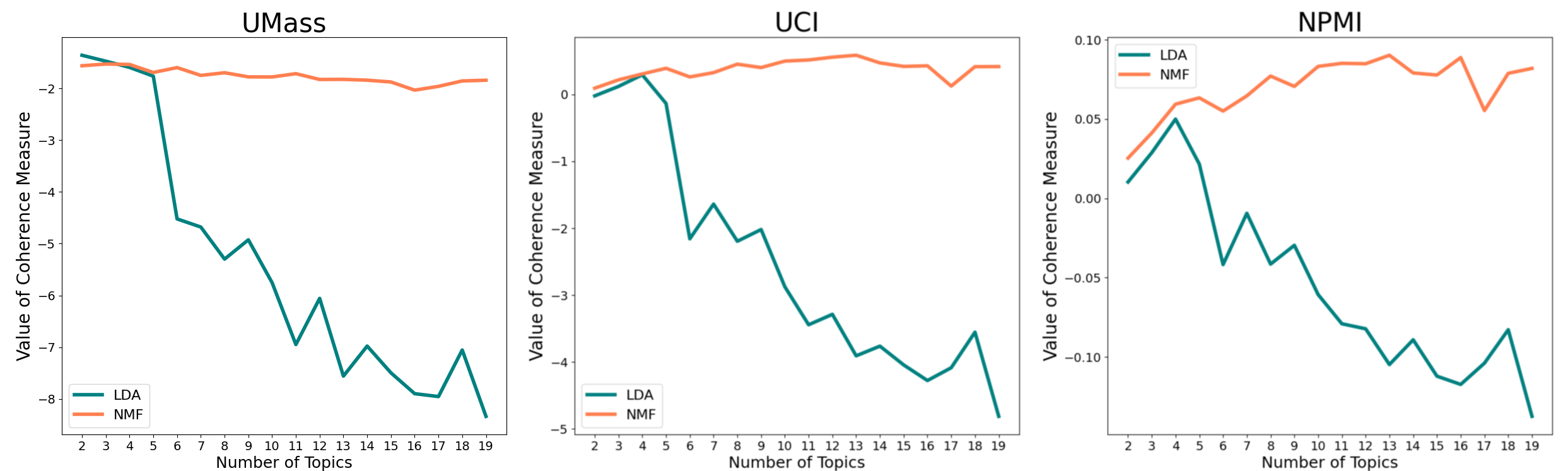}
    \caption{Umass, UCI, and NMPI coherence scores versus the number of topics for LDA (teal) and NMF (orange) models; a higher coherence score is desired. In our study, NMF usually achieves better coherence. For five topics, both models achieve good coherence, and so we further explore the five-topic models}
    \label{fig:coh-graph}
\end{figure}

To assess the topic modeling performance at various values of $t$ (the number of topics), we generated models for between 2 to 19 topics. We leverage various standard coherence measures as  metrics to validate overall model performance and compare models, including UMass, UCI and NPMI coherence scores \cite{roder2015exploring}. Figure \ref{fig:coh-graph} shows that both LDA and NMF models perform very similarly for a small number of topics, but LDA suffers from a drop in coherence across metrics, while NMF does not. This differs from \cite{pandey2017}, where LDA had better UMass coherence than NMF. We note that both LDA and NMF identify topics with similar keywords in many cases. For example, one LDA and one NMF topic both include ``vehicle,'' ``stolen,''  ``victim,'' ``parked,'' ``window'' and ``damage'' in their top ten keywords. 

The coherence scores in with $t=5$ the topics are reasonably coherent for both the NMF and LDA models, and so we pick such for further analysis. Table \ref{tab:nmf-tops} shows key-words representative of each of the five topics learned by the NMF and LDA models, depicting the five most highly weighted key terms that describe the topic and their weights.

\begin{table}
    \begin{center}
    \begin{tabular}{ccc}
    \label{tab:nmf-tops}
    \begin{sc}
    \resizebox{0.6\textwidth}{!}{
    \begin{tabular}{ccc}
    &&\\
    &&\\
    &&\\
    \toprule[1pt]\midrule[0.3pt]
         Topic ID & Key Word & Word Weight \\
         \hline
    NMF0&	driver	&0.024\\
    &	license	&0.016\\
    &	vehicle	&0.014\\
    &	suspended	&0.009\\
    &	traffic	&0.008\\
    \hline
    NMF1&	ms	&0.017\\
    &	mr	&0.005\\
    &	card	&0.005\\
    &	phone	&0.004\\
    &	wallet	&0.004\\
    \hline
    NMF2&	accused	&0.010\\
    &	marijuana	&0.009\\
    &	male	&0.009\\
    &	warrant	&0.008\\
    &	drinking	&0.005\\
    \hline
    NMF3&	vehicle	&0.021\\
    &	parked	&0.011\\
    &	stolen	&0.011\\
    &	window	&0.008\\
    &	damage	&0.007\\
    \hline
    NMF4&	door	&0.012\\
    &	apartment	&0.008\\
    &	home	&0.007\\
    &	entry	&0.004\\
    &	burglary	&0.004\\
    \midrule[0.3pt]\bottomrule[1pt]
    \end{tabular}
    \quad
    \begin{tabular}{ccc}
    &&\\
    &&\\
    &&\\
    \toprule[1pt]\midrule[0.3pt]
         Topic ID & Key Word & Word Weight \\
         \hline
    LDA0&	ms	&0.011\\
    &	victim	&0.004\\
    &	apartment	&0.003\\
    &	mother	&0.003\\
    &	grady	&0.003\\
    \hline
    LDA1&	vehicle	&0.009\\
    &	stolen	&0.005\\
    &	door	&0.005\\
    &	parked	&0.005\\
    &	damage	&0.004\\
    \hline
    LDA2&	license	&0.012\\
    &	warrant	&0.011\\
    &	vehicle	&0.010\\
    &	driver	&0.009\\
    &	suspended	&0.006\\
    \hline
    LDA3&	male	&0.007\\
    &	mr	&0.006\\
    &	marijuana	&0.006\\
    &	suspect	&0.004\\
    &	black	&0.003\\
    \hline
    LDA4&	driver	&0.012\\
    &	card	&0.007\\
    &	store	&0.007\\
    &	account	&0.006\\
    &	vehicle	&0.006\\
    \midrule[0.3pt]\bottomrule[1pt]
    \end{tabular}
    }
    \end{sc}
    \end{tabular}
    \caption{Top five weighted keywords for each learned topic representing a grouping of concepts}
    \end{center}
    \end{table}
    
\subsection{Topic hot-spot identification via kNN-RDE}
\begin{figure}[t]
    \centering
    \includegraphics[width=.35\linewidth]{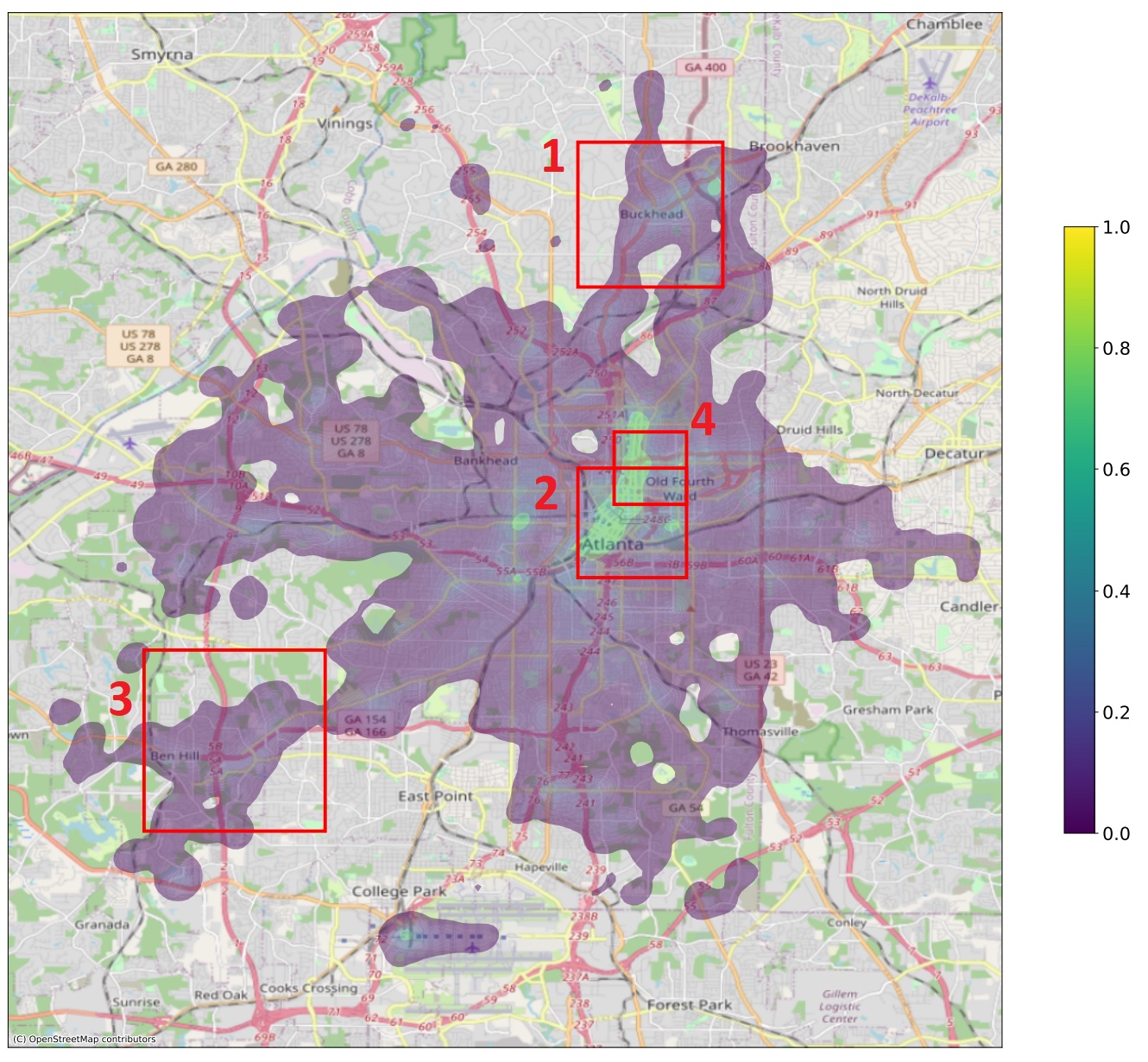}
    \caption{Baseline kernel density estimates (KDE) of the spatial distribution across all incidents and topics. Rectangles indicate further discussed areas.}
    \label{fig:all-inc}
\end{figure}

\begin{figure}
    \centering
    \includegraphics[width= 0.6\linewidth]{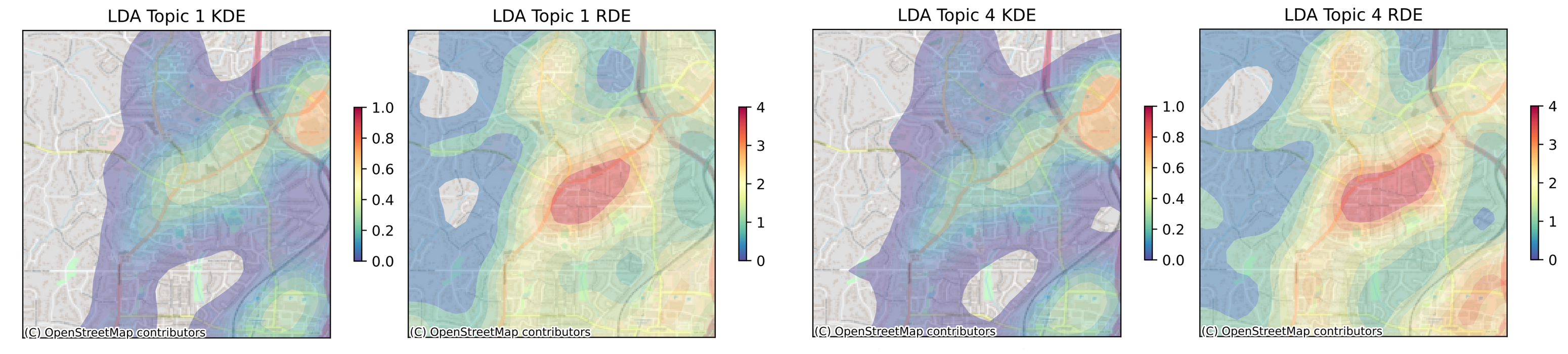}
    \includegraphics[width= 0.6\linewidth]{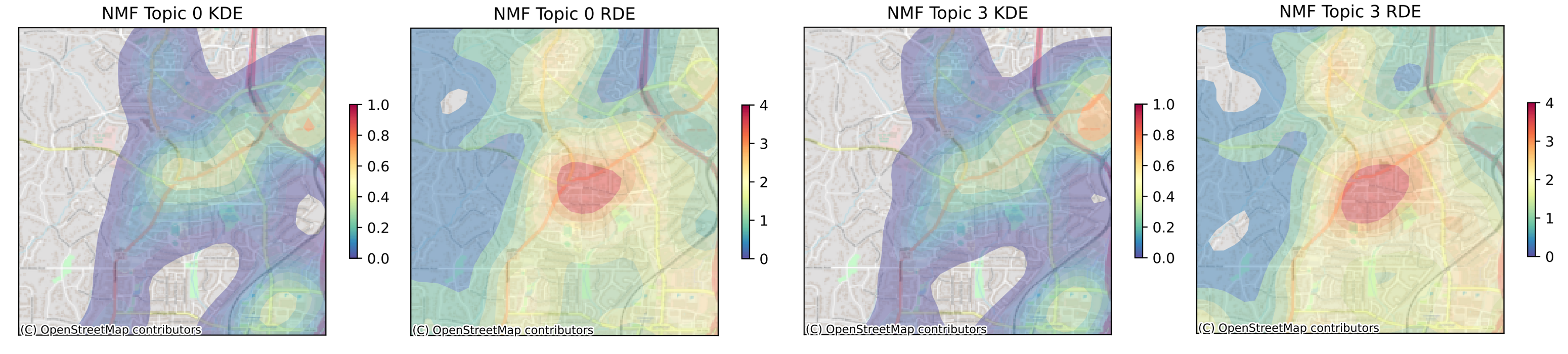}
    \caption{Four similar vehicle-related topics across the two different models, LDA Topic 1 (vehicle, stolen, door, parked, damage), LDA Topic 4 (driver, card, store, account, vehicle) NMF topics, NMF Topic 0 (driver, licence, vehicle, suspended, traffic) and NMF Topic 3 (vehicle, parked, stolen, window damage), are far more prevalent in the Buckhead area of Atlanta (rectangle 1 in Figure \ref{fig:all-inc}). This is especially evident on the kNN-RDE plot, where we can observe incidents in these topics occurring up to four times more often than on average}
    \label{fig:lda-buckhead}
\end{figure}

\begin{figure}
    \centering
    \includegraphics[width= 0.8\linewidth]{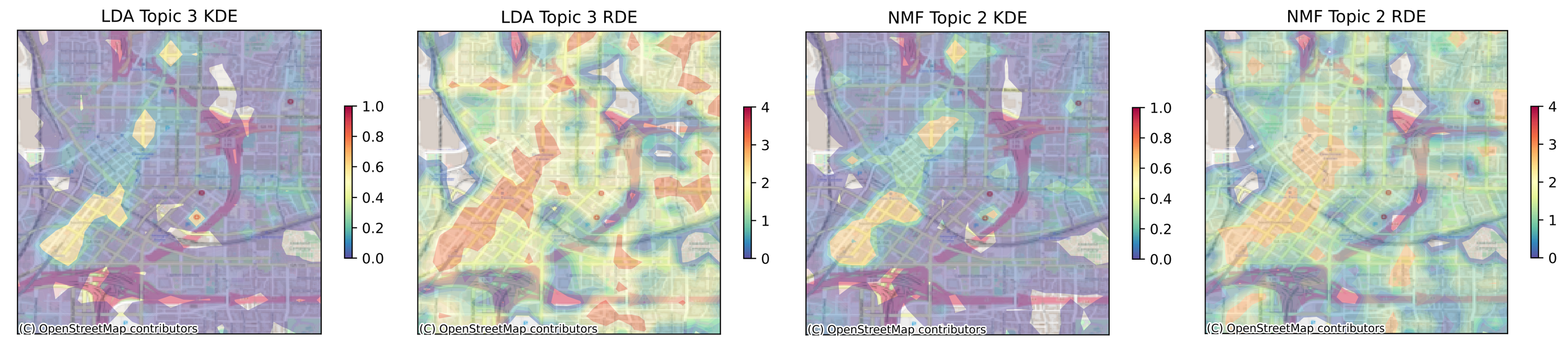}
    \caption{Two similar topics, LDA Topic 3 (male, mr, marijuana, suspect, black) and NMF Topic 2 (accused, marijuana, male, warrant, drinking), are the most prevalent in the downtown area (rectangle 2 in Figure \ref{fig:all-inc}), indicating that drug and alcohol related incidents are frequent here. Note there are more hotspots and more drastic variation on the kNN RDE plots.}
    \label{fig:downtown}
\end{figure}

\begin{figure}
    \centering
    \includegraphics[width=0.8\linewidth]{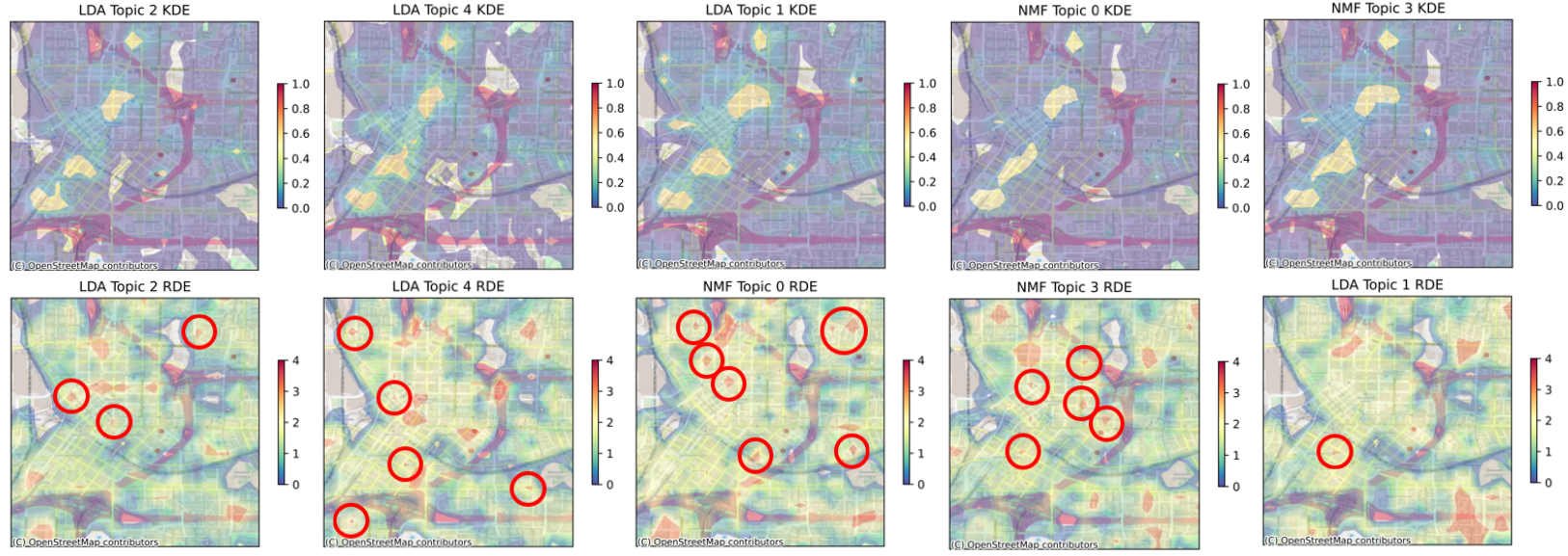}
    \caption{Four similar vehicle-related topics (LDA Topic 2, LDA Topic 4, NMF Topic 0 and NMF Topic 3) show small, local hotspots in the downtown area (rectangle 2 in \ref{fig:all-inc}), at intersections that are particularly dangerous or where traffic stops often occur, while LDA Topic 1 (vehicle, stolen, door, parked, damage) and NMF Topic 3 show a larger hotspot in the north-central area of downtown. The small hotspots circled in red on the RDE plots are so fine-grained that we can identify specific blocks and intersections where these topics occur more often.}
    \label{fig:downtown-cars}
\end{figure}
\begin{figure}
    \centering
    \includegraphics[width= 0.8\linewidth]{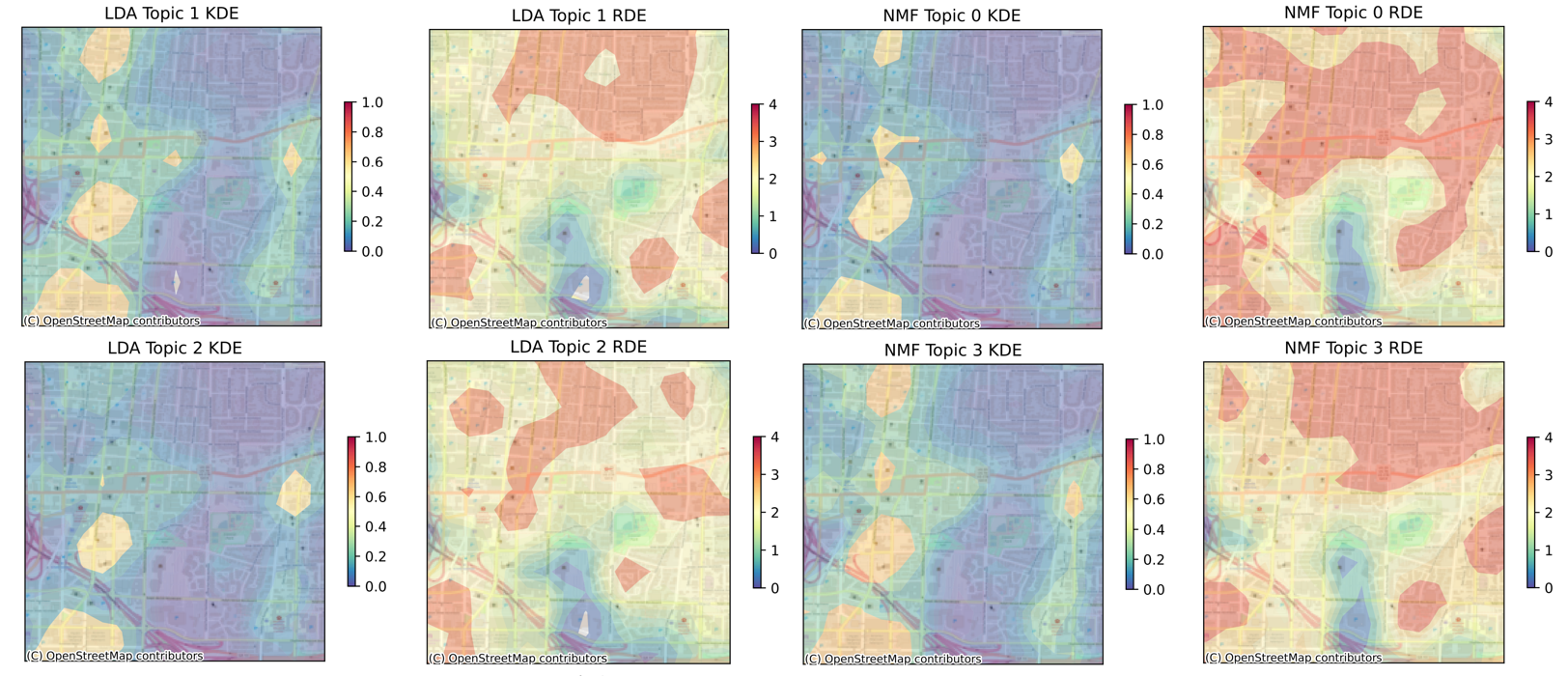}
    \caption{Four similar vehicle-related topics, LDA Topic 1 (vehicle, stolen, door, parked, damage), LDA Topic 2 (license, warrant, vehicle, driver, suspended), NMF Topic 0 (driver, license, vehicle, suspended, traffic) and NMF Topic 3 (vehicle, parked, stolen, window, damage), have similar hot-spots in the midtown area (rectangle 4 in Figure \ref{fig:all-inc}).}
    \label{fig:midtown1}
\end{figure}

In this section, we examine the spatial distribution of the LDA and NMF topics using the kNN relative density estimate (kNN-RDE). The following figures and results for the kNN-RDE were generated by defining a 1000 $\times$ 1000 grid over the city and calculating the $\hat{g}_m$ relative density estimates (with $k=250$) as described in Section \ref{sec:knn-relative-density}. As a baseline, we depict plotted the density of all incidents in Figure \ref{fig:all-inc}. There are naturally high concentrations in areas such as downtown and midtown (boxes 2 and 4), along with various major roadways, highlighting the importance of relative estimation. Using kNN-RDE allows for the identification of topic trends relative to this background within differing neighborhoods.

We identify areas in the city where certain topics occurred more or less frequently and to detect hot-spots for a given topic. Using the kNN-RDE, we can clearly identify some interesting trends in certain neighborhoods in the city. We will also show KDE plots for each topic that are generated using all incidents that are  about a given topic where each incident is equally weighted. 

As an example, if we look at vehicle-related topics in Buckhead (Figure \ref{fig:lda-buckhead}), a relatively affluent area,  we can see that there are very clear hot-spots in LDA topics 1 and 4, which both relate to vehicles and occur up to four times more often in this area than in the city overall. This is also evident in NMF topics 0 and 3, which also include the key word "vehicle". The values of the relative density show how many times more or less often the incident, which is behavior hidden on the KDE plot, which only depicts the absolute density of events and thus obscures the trend.

We can also look at the downtown area. There are several notable topic hotspots in this area, especially for LDA Topic 3 (male, mr, marijuana, suspect, black) and NMF Topic 2 (accused, marijuana, male, warrant, drinking). We can reasonably infer that crimes relating to drugs and alcohol are more common in this area, which is a busy area with bars, hotels, shops and nightlife. 

Vehicle-related topics have very localized hotspots in this area as shown in Figure \ref{fig:downtown-cars}. These local hotspots may show particularly dangerous cross streets or areas where traffic stops often occur. For example LDA Topic 2 (license, warrant, vehicle, driver, suspended), LDA Topic 4 (driver, card, store, account, vehicle), NMF Topic 0 (driver, license, vehicle, suspended, traffic) and NMF Topic 3 (vehicle, parked, stolen, window, damage) show this pattern. By contrast, KDE completely misses these local hotspots. LDA Topic 1 (vehicle, stolen, door, parked, damage) does not show a scattering of small hotspots, but it does show one slightly larger hotspot in the central downtown area, where we also see a slightly larger hotspot for NMF topic 3. This may indicate that traffic violations occur at specific intersections more often, but other types of vehicle damage or theft tend to cluster in this location, where there are many tourist attractions and hotels. 

Vehicle-related topics are visible in midtown Atlanta (Figure \ref{fig:midtown1}). We will also use this area to compare a kernel density estimate (KDE) plot with the kNN-RDE plots and examine some key benefits of the kNN-RDE approach.  LDA Topic 1 (vehicle, stolen, door, parked, damage) and NMF Topic 3 (vehicle, parked, stolen, window, damage) indicate that these areas have a high number of incidents relating to vehicular damage or theft. Since the hotspots for LDA Topic 3 are very clear in this area, we will also use this topic and area to compare the KDE for incident density to the kNN-RDE for incident relative density. We can then see in Figure \ref{fig:midtown1} that both of these methods give a less nuanced view of the incident hotspots than the kNN-RDE for relative density. For the first method, we can see that only three hotspots were detected, and for the second, we can observe five total. The hotspots found with the kNN-RDE are partially completely obscured in this view, especially the ones in the northwest corner of the plot. Since the incident density estimate shows only how many incidents are expected in this area (based on how many incidents are in the area in the data), we cannot get as fine-grained a view as when we look at the relative density.

\begin{figure}
    \centering
    \includegraphics[width= 0.8\linewidth]{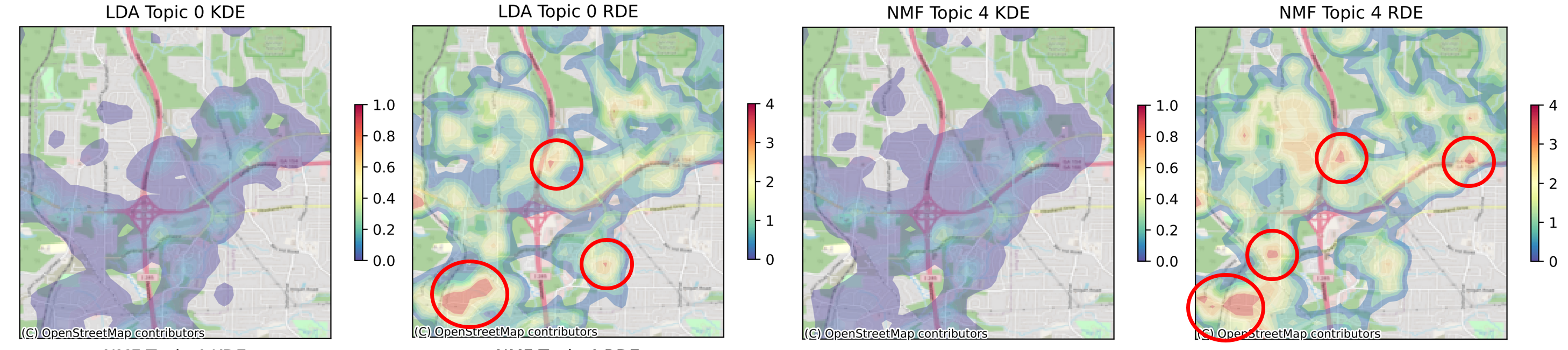}
    \caption{Two similar topics across the different models (for Box 3 in Figure \ref{fig:all-inc}). The top-five keywords from the LDA Topic are (ms, victim, apartment, mother, grady) and for the NMF Topic (door, apartment, home, entry, burglary). The hotspots identified by kNN-RDE (in red) are missed by KDE.}
    \label{fig:benhill}
\end{figure}

The Ben Hill area of Atlanta (Box 3 in Figure \ref{fig:all-inc}) likewise demonstrates the importance of relative estimation under the context of low population density. In this area, we observe clear hotspots for LDA Topic 0 (ms, victim, apartment, mother, grady) and NMF Topic 4 (door, apartment, home, entry, burglary), as seen in Figure \ref{fig:benhill}. Keywords, such as ``apartment'' and ``home'' for both LDA Topic 0 and NMF Topic 4, indicate the type of incidents that take place in residences.

\section{Conclusion}
We leverage automatic topic identification from a large corpus of events and relative estimation to identify trends and true hot-spots beyond agglomerations attributable to external factors. The use of topic modeling allows the identification of wider trends that may be unobserved by police. Relative density estimation identifies key trends which are lost in ordinary KDE, since the underlying distribution of events is highly nonuniform.
Our method of automatically identifying topics and identifying hotspots is readily adaptable to domains such as fraud detection or trend identification in geographically tagged customer service records and textual reviews or generated reports in other service settings.

\section*{Acknowledgement}
    This work is partially supported by an NSF CAREER CCF-1650913, and NSF DMS-2134037, CMMI-2015787, CMMI-2112533, DMS-1938106, DMS-1830210, the Atlanta Police Foundation, and the Coca-Cola Foundation. The authors would also like to thank the support and expert knowledge input from the Atlanta Police Department.

\bibliography{main.bib}
\end{document}